# Distance-based mutual congestion feature selection with genetic algorithm for high-dimensional medical datasets

Hossein Nematzadeh[1]*, Joseph Mani[1], Zahra Nematzadeh[2], Ebrahim Akbari[3], Radziah Mohamad[4]

[1]Department of Mathematics and Computer Science, Modern College of Business and Science, Muscat, Oman
[2]Otolaryngology Department, School of Medicine, Emory University, Atlanta, Georgia, 30322, USA
[3]Department of Computer Engineering, Sari Branch, Islamic Azad University, Sari, Iran
[4]Faculty of Computing, Universiti Teknologi Malaysia, 81310 Skudai, Malaysia
* The corresponding author is available via hossein.nematzadeh@mcbs.edu.om or hn_61@yahoo.com

## Abstract

Feature selection poses a challenge in small-sample high-dimensional datasets, where the number of features exceeds the number of observations, as seen in microarray, gene expression, and medical datasets. There isn't a universally optimal feature selection method applicable to any data distribution, and as a result, the literature consistently endeavors to address this issue. One recent approach in feature selection is termed frequency-based feature selection. However, existing methods in this domain tend to overlook feature values, focusing solely on the distribution in the response variable. In response, this paper introduces the Distance-based Mutual Congestion (DMC) as a filter method that considers both the feature values and the distribution of observations in the response variable. DMC sorts the features of datasets, and the top 5% are retained and clustered by KMeans to mitigate multicollinearity. This is achieved by randomly selecting one feature from each cluster. The selected features form the feature space, and the search space for the Genetic Algorithm with Adaptive Rates (GAwAR) will be approximated using this feature space. GAwAR approximates the combination of the top 10 features that maximizes prediction accuracy within a wrapper scheme. To prevent premature convergence, GAwAR adaptively updates the crossover and mutation rates. The hybrid DMC-GAwAR is applicable to binary classification datasets, and experimental results demonstrate its superiority over some recent works. The implementation and corresponding data are available at https://github.com/hnematzadeh/DMC-GAwAR

## 1. Introduction

Feature selection in small-sample high-dimensional datasets [1] plays a crucial role, offering three main advantages. Firstly, it enhances the accuracy of prediction models. Secondly, it reduces the model training execution time. Thirdly, it aids decision-makers in identifying important features. In the context of biological or medical high-dimensional data, these datasets may take the form of microarrays or gene expression formats. The feature values can either be distinct integers within a discrete set or any real numbers within a specified range. These datasets often encompass a vast number of features with very few samples, resulting in high dimensionality. The literature contains numerous works [2-6] that address the challenge of feature selection for such datasets [7]. One of the major categories of feature selection methods is called filter feature selection method. While filter feature selection methods may encompass a machine learning model in the feature scoring process, the trained model is not commonly employed for testing or prediction. Instead, statistical, or mathematical measures are typically utilized to assess the relevance of features, often by evaluating their correlation with the target variable or employing another scoring method. The primary focus is on ranking or scoring features based on certain criteria, and the selected features are then used independently in subsequent steps of the analysis or model training. Once the features are sorted based on the selected criteria (e.g., correlation coefficients, information gain), the top features are typically chosen according to a predetermined threshold or a specified number of features to be retained. Filter methods are computationally efficient and can be applied independently of a specific learning algorithm.

A new approach in filter feature selection has emerged recently, known as frequency-based feature selection [8], which incorporates various methods for solving classification problems. Typically, these methods involve sorting the features within datasets and assigning a score to each feature based on the reordered response variable. The current frequency-based feature selection methods determine the score for each feature by analyzing the frequency of non-separable elements in mutual congestion region [8-11] or identifying the maximum pattern recognized in the response variable [12]. These methods, however, overlook the actual values of features when calculating scores. As a result, the primary hypothesis of this paper posits that taking feature values into account may yield a positive impact, allowing for more realistic weights assigned to features. This, in turn, could enhance the accuracy of the filter frequency-based ranking, particularly in specific scenarios.

The other major category of feature selection is called wrapper feature selection where the selection of features is treated as a search problem, and different subsets of features are evaluated using a predictive model [13]. The process involves training and evaluating the model's performance with different feature subsets. Wrapper methods use a specific machine learning model's performance as a criterion to guide the selection of features. These methods can be computationally expensive because they involve training and testing the model for various feature subsets, but they often result in more accurate feature selection tailored to the chosen model. To formulate a search strategy for wrapper feature selection, one effective approach is the utilization of evolutionary algorithms [12, 13], such as the Genetic Algorithm (GA) [14]. The second hypothesis posited in this paper asserts that adaptive crossover and mutation rates may mitigate premature convergence, thereby enhancing the optimization process of GA. Grounded in these two hypotheses, we enumerate the contributions of this paper as follows:

1- A new frequency-based filter, namely, Distance-based Mutual Congestion (DMC).
2- A wrapper feature selection method, namely, Genetic Algorithm with Adaptive Rates (GAwAR).

    - GAwAR optimizes in a specific search space generated by the top 5 percent features recognized by DMC clustered into 100 clusters. The aim of clustering here is to decrease the possible multicollinearity among selected features by DMC. GAwAR also uses adaptive crossover and mutation rates to prevent premature convergence.

3- A hybrid feature selection method, namely, DMC-GAwAR

The structure of the remainder of this paper is outlined below. Section 2 delves into related works, identifying existing gaps that serve as inspiration for our contributions. In Section 3, we provide a succinct overview of the Genetic Algorithm, (GA) explaining why it has been chosen to approximate the optimized feature subset in this study. Section 4 introduces our proposed method, Distance-based Mutual Congestion Genetic Algorithm with Adaptive Rates (DMC-GAwAR). Section 5 presents the experimental results, and, finally, Section 6 concludes the paper.

## 2. Related works

Given that the primary contribution of this paper is the introduction of a new frequency-based ranker, our initial focus centers on examining existing frequency-based methods to identify gaps that serve as inspiration for the development of Distance-based Mutual Congestion (DMC) in Section 2.1. Subsequently, certain recent feature selection methods specifically designed for classification problems are introduced in Section 2.2.

### 2.1 Frequency-based rankers

All the existing frequency-based rankers, as outlined in Table 1, initially arrange the feature values in ascending order. Subsequently, each ranker employs a distinct technique for feature ranking based on the reordered response variable. Some of these methods are designed for application in multi-class datasets, while others are exclusively suited for binary datasets. Depending on the class constraint, the time complexities of existing frequency-based methods can be classified into two groups: those with time complexities of $O(mn\,log\,n)$ and others with time complexity of $O(mn\,log\,n\,+\,mln)$, where m, n, and l represent the number of features, rows, and labels (classes) in the respective dataset. The frequency-based rankers outlined in Table 1 do not consider the feature values once the features are sorted. Their formulation depends solely on the reordered response variable to assign a score to each feature.

Recognizing this limitation in frequency-based rankers, we propose that incorporating the ascendingly sorted feature values, in addition to the reordered response variable, has the potential to enhance rankings. This enhancement stems from the idea of assigning more realistic scores to the features. This forms the basis for the development of the Distance-based Mutual Congestion (DMC) filter.

Table 1. Existing frequency-based methods

| Frequency-based ranker | Applicability | Time complexity | Year |
|---|---|---|---|
| Mutual Congestion [8] | Binary | $O\ (mn\,log\,n)$ | 2019 |
| Sorted Label Interference [9] | Binary | $O\ (mn\,log\,n)$ | 2021 |
| Sorted Label Interference-$\gamma$ [10] | Binary | $O\ (mn\,log\,n)$ | 2022 |
| Extended Mutual Congestion [11] | Multiclass | $O\ (mn\,log\,n\,+\,mln)$ | 2022 |
| Maximum Pattern Recognition [12] | Multiclass | $O\ (mn\,log\,n)$ | 2024 |

## 2.2 Related feature selection methods

Numerous hybrid methods exist, with one notable approach involving the integration of both filter and wrapper techniques. In this context, the hybrid method combines the advantages of a filter ranker with a wrapper method, allowing the selection of the top $k$ features determined by the filter ranker. This approach is part of the broader landscape of hybrid feature selection methods aiming to leverage the strengths of different selection strategies for enhanced performance. In the realm of hybrid feature selection methods, all frequency-based rankers in Table 1 have been employed as filter methods within hybrid frameworks, with the exception of Sorted Label Interference (SLI) [9] which used within an ensemble feature selection method. Ensemble feature selection instead of relying on a single feature selection algorithm, leverage the diversity of multiple algorithms to collectively identify the most informative and relevant features. Another example of ensemble feature selection approach is Pareto-based Ensemble of Feature Selection (PEFS) [15] as an ensemble method of four filters. Mutual Congestion (MC) was utilized in conjunction with the filter Whale Optimization Algorithm (WOA) to form the hybrid WOA-MC [8]. Sorted Label Interference-γ (SLI-γ) was fused with the wrapper Genetic Algorithm (GA) to establish the hybrid $GA_{rank\&rand}$ [10]. Extended Mutual Congestion (EMC) was integrated with the wrapper Discrete Weighted Evolution Strategy (DWES), and Maximum Pattern Recognition (MPR) was combined with Multi-objective Discrete Evolution Strategy (MDES), resulting in the development of EMC-DWES [11] and MPR-MDES [12], respectively. All the previously mentioned methods underwent evaluation using benchmark high-dimensional microarray medical datasets. Each method exhibited specific strengths and weaknesses, accompanied by certain limitations. For instance, WOA-MC and $GA_{rank\&rand}$ were exclusively suitable for classification datasets featuring a binary response variable. This limitation may be attributed to the binary nature of the rankers in MC and SLI-γ, respectively.

Examining the literature uncovers additional hybrid approaches, including EIT-bBOA [16], MGWO [17], HyCluster [18], and Hybrid filter-genetic feature selection [19], each contributing unique perspectives to the field. Notably, these methods did not incorporate frequency-based rankers. Ensemble Information Theory-based binary Butterfly Optimization Algorithm (EIT-bBOA) [16], an innovative hybrid approach integrating a filter, a wrapper, and an ensemble method. In its methodology, EIT-bBOA incorporates the Minimal Redundancy-Maximal New Classification Information (MR-MNCI) as a filtering technique, effectively eliminating 80% of non-relevant features. The Information Gain binary Butterfly Optimization Algorithm (IG-bBOA) is then employed for optimal feature subset selection. The final step involves a manual selection of the top 30 features using an ensemble method combining ReliefF and Fisher score.

The Modified Gray Wolf Optimization (MGWO) [17] cleverly integrated both filters and wrappers to perform feature selection. In its approach, MGWO employed filters such as the ReliefF algorithm and Copula entropy to initially rank features. The primary objective was to diminish the search space in addressing large-scale feature selection challenges, leveraging correlation measures to prevent the inclusion of low-quality features in the initial population. Furthermore, the differential evolution algorithm was incorporated to extend the exploration of the standard Gray Wolf Optimization (GWO) search space. The outcomes demonstrated the effectiveness of MGWO, showcasing commendable results across 10 gene expression datasets.

HyCluster [18] is a hybrid feature selection scheme designed for high-dimensional data, addressing time complexity concerns through the incorporation of feature clustering. The HyCluster process unfolds in two stages: initially, features are ranked based on their relevance to the class label, utilizing the Symmetrical Uncertainty (SU) index. The top $\theta$ features (where $\theta$ is set to $\sqrt{n} \times \ log\, n$, and $n$ represents the number of features) are selected using a filter approach. Subsequently, three clustering techniques—Kmeans, shared nearest neighbor, and spectral clustering—are applied to cluster the remaining top θ features from the filter stage. The primary objective of clustering is to segment the search space of the problem effectively. In the wrapper stage, an Incremental Wrapper Subset Selection search strategy (IWSS) is executed on the best clusters, leading to the final selection of features from each cluster. HyCluster's performance was assessed across four facial image datasets and four microarray datasets, demonstrating superior results in terms of accuracy and computational complexity compared to counterparts in existing works.

Hybrid filter-genetic feature selection [19] utilized information gain, information gain ratio, and Chi-squared initially to rank the features of the dataset with the aim of keeping only the top 5% of the features. Then Genetic Algorithm (GA) was used within a wrapper scheme for optimization and finding the subset of features that maximizes the classification accuracy. The termination criteria of the proposed GA were either obtaining satisfactorily optimal fitness or reaching maximum generations.

As a result, this paper also posits the hypothesis that hybrid feature selection methods, incorporating both filter and wrapper techniques, could significantly enhance the classification accuracy of prediction models within a reasonable computational time. Consequently, the study explores the integration of the Distance-based Mutual Congestion (DMC) as a novel frequency-based ranker and the proposed Genetic Algorithm with Adaptive Rates (GAwAR) on high-dimensional medical datasets.

## 3. Application of GA on high-dimensional feature selection

Genetic Algorithm (GA) [14] offers a robust and effective alternative for feature selection in high-dimensional datasets. These algorithms draw inspiration from natural selection and evolution, creating a population of potential solutions that undergo iterative generations of crossover, mutation, and selection processes. This approach allows genetic algorithms to explore a vast solution space efficiently. In the context of high-dimensional datasets, where the number of features is large, genetic algorithms excel in searching for optimal feature subsets. They can handle complex and nonlinear relationships among features, providing a versatile solution for capturing the most relevant information. By evaluating and evolving potential feature subsets over multiple generations, GA can adapt to the intricacies of high-dimensional data, helping to identify informative features while mitigating the curse of dimensionality. The ability of GA to balance exploitation (via crossover) and exploration (via mutation) makes them suitable for optimizing feature selection in a way that enhances classification accuracy and generalization performance. Overall, genetic algorithms stand out as a powerful and adaptive tool for feature selection in the face of high-dimensional datasets [10, 20].

Nevertheless, the majority of GA methods typically employ fixed values for crossover and mutation rates. In this paper, we introduce an innovative approach by making these rates adaptive, aiming to facilitate the GA in escaping potential local maxima. This adaptability enhances the algorithm's ability to navigate the solution space dynamically, potentially leading to more effective exploration and discovery of optimal solutions.

## 4. Proposed method

The pipeline for the proposed method (DMC-GAwAR) is illustrated in Figure 1. First, the features of the dataset are scored and subsequently ranked using the proposed Distance-based Mutual Congestion (DMC) filter. Thereafter, the top 5% of features are selected for the next stage. However, to address multicollinearity and enhance the diversity of selected features, the top 5% are clustered into 100 groups using the KMeans clustering algorithm before applying the proposed Genetic Algorithm with Adaptive Rates (GAwAR). From each cluster, one feature is randomly selected. Therefore, GAwAR will use a feature space of 100 minimally correlated features to form a problem-specific search space. Finally, GAwAR is employed as a wrapper approach to select the best 10 features, collectively improving the prediction accuracy over the initial set of 100 individual features. The proposed GA balances between exploitation and exploration by dynamically adapting crossover and mutation rates.

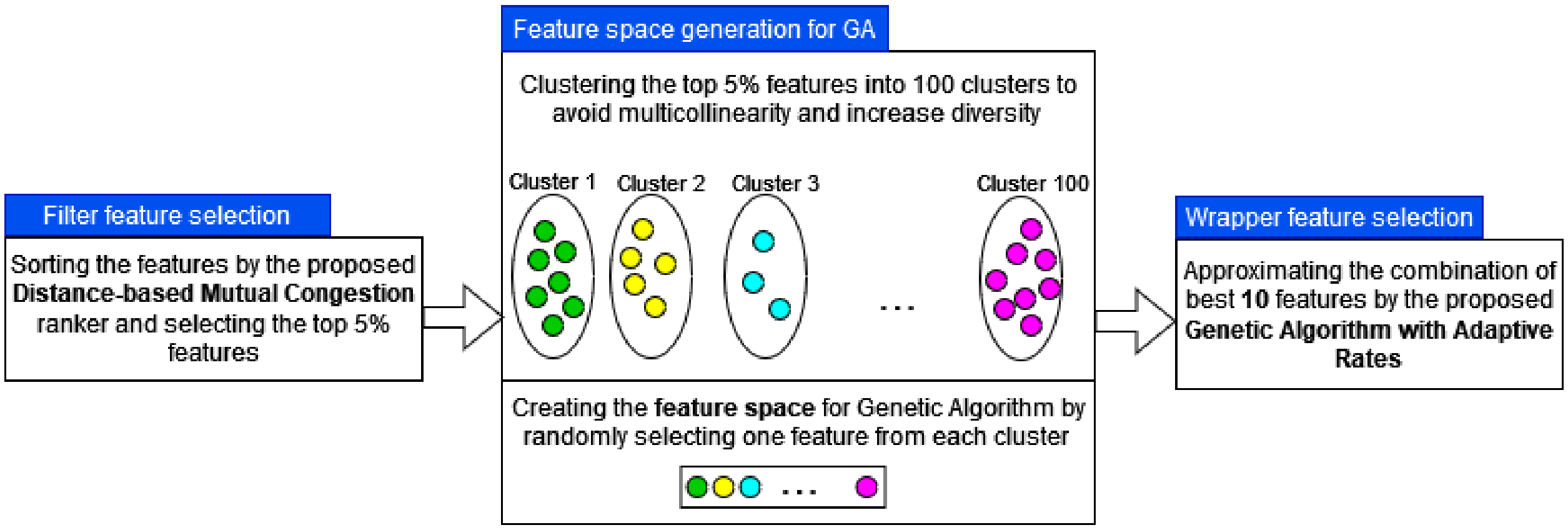

Figure 1. General view of DMC-GAwAR

### 4.1 Distance-based Mutual Congestion

Assuming the high-dimensional binary classification dataset D consists of $n$ records and $m$ features, as illustrated in Table 2. The records and features of the dataset can be represented as sets $R = (r_1, r_2, r_3, ..., r_n)$ and $F = (f_1, f_2, f_3, \ldots, f_m)$, respectively in which $m \gg n$. Given that the binary classification dataset inherently consists of two classes, they can be represented by the set $C = (c_1, c_2)$. Consequently, $c_i$ in Table 2 corresponds to either $c_1$ or $c_2$. According to the given setup, each feature $f_i$ can be represented as a vector across all samples using Eq.1 and subsequently each record (sample) $r_i$ can be shown as a vector with feature notations using Eq.2. As a result, the dataset D can also be represented as $D = \{f_{ij}\}$, where $1 \leq i \leq n$ and $1 \leq j \leq m$.

$$f_j = (f_{1j}, f_{2j}, f_{3j}, \dots, f_{nj}) \qquad j = 1,2, \dots, m \qquad (1)$$

$$r_i = (f_{i1}, f_{i2}, f_{i3}, \dots, f_{im}) \qquad i = 1,2, \dots, n \qquad (2)$$

We can also define the set of all records of a same class as $y^{C_i}$.

$$y^{C_i} = \{ {r_j}^{C_i} \mid j = 1,2, \dots n_i \},\ n_i = |y^{C_i}|,\ i \in \{1,2\} \qquad (3)$$

Table 2. Schematic of a high-dimensional dataset for binary classification

| | $f_1$ | $f_2$ | $f_3$ | $\cdots$ | $f_{m-1}$ | $f_m$ | $c_i$: $i \in \{1,2\}$ |
|---|---|---|---|---|---|---|---|
| $r_1$ | $f_{11}$ | $f_{12}$ | $f_{13}$ | $\cdots$ | $f_{1m-1}$ | $f_{1m}$ | $c_i$ |
| $r_2$ | $f_{21}$ | $f_{22}$ | $f_{23}$ | $\cdots$ | $f_{2m-1}$ | $f_{2m}$ | $c_i$ |
| $r_3$ | $f_{31}$ | $f_{32}$ | $f_{33}$ | $\cdots$ | $f_{3m-1}$ | $f_{3m}$ | $c_i$ |
| $\vdots$ | $\vdots$ | $\vdots$ | $\vdots$ | $\vdots$ | $\vdots$ | $\vdots$ | $\vdots$ |
| $r_n$ | $f_{n1}$ | $f_{n2}$ | $f_{n3}$ | $\cdots$ | $f_{nm-1}$ | $f_{nm}$ | $c_i$ |

In the domain of frequency-based feature selection, a fundamental preprocessing step involves the sorting of feature values, aligning them in a structured order. Simultaneously, the response variable undergoes sorting to maintain coherence. This initial stage serves as a pivotal preparatory measure in any frequency-based method. Notably, the Distance-based Mutual Congestion (DMC) methodology adheres to this rule, further emphasizing the significance of sorted data in the context of feature selection methodologies. Distance-based Mutual Congestion (DMC) is originally developed for a binary classification dataset.

For simplicity, let's assume the two classes are represented as $C_1$ and $C_2$. Additionally, let's assume that the values of feature $f_1$ have been sorted in ascending order, and the response variable is correspondingly affected by the same sorting in Figure 2. Likewise, assume that in our example the dataset has 16 samples. And X and Y denote the set of observations with $C_1$ and $C_2$ labels. We can define the following based on Figure 2 assuming the sorted response variable starts with $C_1$ and ends with $C_2$. It is noteworthy to mention that mutual congestion region refers to the interference region in which labels are not separable. For example, it starts from the first appearance of $C_2$ until the last appearance of $C_1$ in Figure 2.

$X_{NMC}$ = the set of feature values with consecutive $C_1$ label in the response variable which are not in mutual congestion region = 1.8, 2.3, 2.4, 2,45, 2.9 $Y_{NMC}$ = the set of feature values with consecutive $C_2$ label in the response variable which are not in mutual congestion region = 4, 4.2, 5.2, 5.5, 5.9

$X_{MC}$ = the set of feature values with $C_1$ label in mutual congestion region = 3.15, 3.25, 3.3

$Y_{MC}$ = the set of feature values with $C_2$ label in mutual congestion region = 3, 3.1, 3.2

$Y_{min}$ = the corresponding feature value of the first appearance of label $C_2$ = 3

$X_{max}$ = the corresponding feature value of the last appearance of label $C_1$ = 3.3

As such, DMC sorts the features based on minimization of Eq. 4 . The DMC in Eq.4 corresponds to its equivalent expression in Eq.5.

$$DMC = \frac{|X_{MC} - Y_{min}|}{|X_{NMC} - Y_{min}|} + \frac{|Y_{MC} - X_{max}|}{|Y_{NMC} - X_{max}|} \quad (4)$$

$$DMC = \frac{\sum |x_i{}^{MC} - Y_{min}|}{\sum |x_j{}^{NMC} - Y_{min}|} + \frac{\sum |y_l{}^{MC} - X_{max}|}{\sum |y_k{}^{NMC} - X_{max}|} \quad (5)$$

where

$$\forall x_i{}^{MC} \in X_{MC}, \forall x_j{}^{NMC} \in X_{NMC}, \forall y_l{}^{MC} \in Y_{MC}, \forall y_k{}^{NMC} \in Y_{NMC}$$

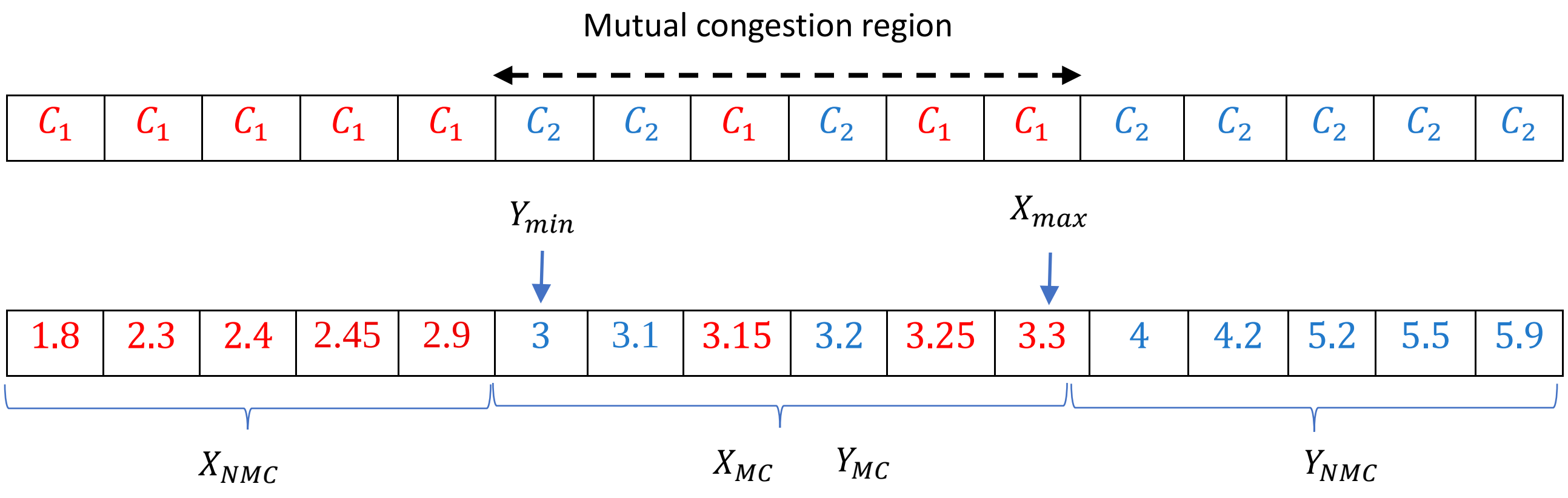

Figure 2. Illustrating the DMC procedure with a sample of 16 observations. The feature values are sorted at the bottom, leading to a reordering of the response variable according to the sorted feature values at the top.

According to Eq.5 the value of DMC for the feature in Figure 2 is calculated as follows.

$$\sum |x_i^{MC} - Y_{min}| = |3.15-3| + |3.25-3| + |3.3-3| = 0.15 + 0.25 + 0.3 = 0.7$$

$$\sum |x_j^{NMC} - Y_{min}| = |1.8-3| + |2.3-3| + |2.4-3| + |2.45-3| + |2.9-3|$$
$$= 1.2 + 0.7 + 0.6 + 0.55 + 0.1 = 3.15$$

$$\sum |y_l^{MC} - X_{max}| = |3-3.3| + |3.1-3.3| + |3.2-3.3| = 0.3 + 0.2 + 0.1 = 0.6$$

$$\sum |y_k^{NMC} - X_{max}| = |4-3.3| + |4.2-3.3| + |5.2-3.3| + |5.5-3.3| + |5.9-3.3|$$
$$= 0.7 + 0.9 + 1.9 + 2.2 + 2.6 = 8.3$$

$$DMC = \frac{0.7}{3.15} + \frac{0.6}{8.3} = 0.22 + 0.07 = 0.29$$

The DMC for the feature in Figure 2 equals 0.29. Since DMC is a minimization formula, a lower value is preferable. Consequently, we can generally infer that a lower DMC value suggests that the corresponding feature in Figure 2 exhibits good separability power. In an ideal scenario where mutual congestion region does not contain any observation, the DMC value equals zero. However, in the worst case, the DMC value is a positive value dependent on the feature values. As evident, DMC calculates the separability power of the feature based on the feature values corresponding to the sorted response variable. This approach differs from the original MC measure, which solely exploits the reordered response variable and does not consider the corresponding feature values. For instance, the separability power of the feature in Figure 2 calculated by MC is $\frac{3+3}{5+3+3+5} = \frac{6}{16} = 0.38$. The reason for this is that MC divides the number of samples in the mutual congestion region (3 samples with $C_2$ label and 3 samples with $C_1$label) by the total number of samples either inside or outside the mutual congestion region. This includes 5 samples outside the mutual congestion region with the $C_1$ label, 3 samples with the $C_2$ label, and 3 samples with the $C_1$ label, all inside the mutual congestion region, as well as 5 samples outside the mutual congestion region with the $C_2$ label.

Clearly, DMC assigns a more favorable weight to the corresponding feature in Figure 2 confirming that DMC theoretically outperforms MC in this specific example in Figure 2. Additionally, since data exists in different shapes, we should also examine this experimentally, which will be done in Section 5. In summary, the proposed DMC effectively organizes the dataset by sorting its features, retaining only the top 5% of the most informative ones for subsequent stages. This strategy not only streamlines the dataset but also improves efficiency, emphasizing the extraction of key information.

4.2 Feature space generation

Handling high dimensional datasets, which involve a large number of features, poses a challenge. Even by retaining only the top 5% of features, the dataset can remain significantly large. Moreover, there's a possibility of multicollinearity among these top features, where several independent features are correlated. To address these, we take additional steps to both eliminate possible multicollinearity and further reduce the search space for the Genetic Algorithm (GA). Specifically, the top 5% features obtained from the filter Distance-based Mutual Congestion (DMC) ranker are subjected to clustering using the KMeans clustering algorithm. KMeans clustering is chosen for its speed and efficiency in clustering. A notable advantage is that it doesn't demand specialized hyperparameter tuning, in contrast to existing clustering methods such as hierarchical clustering, where considerations about the linkage type are essential. While KMeans is relatively straightforward, it does require specifying the number of clusters (q) in advance. In our proposal, we fix this hyperparameter to q=100. As a result, the set of selected top 5% features by DMC which is denoted by Eq.5 is further clustered into $q$ clusters by KMeans clustering algorithm in Eq.6 where $c_1, c_2, \dots, c_q$ are the corresponding clusters.

$$F^{DMC} = \left( {f_1}^{DMC}, {f_2}^{DMC}, {f_3}^{DMC}, \dots, {f_{m'}}^{DMC} \right), \quad m' = 0.05 \times m \tag{5}$$

$$C^q = KMeans(F^{DMC}, q) = c_1, c_2, \dots, c_q \tag{6}$$

The idea is that cluster members should theoretically have common characteristics. Instead of considering all top 5% features in $F^{DMC}$, we can select one feature from each cluster of $C^q$ randomly, precisely selecting $q$ features to generate the feature space for GA. Genetic Algorithm with Adaptive Rates (GAwAR) will utilize these $q$ features to approximate the best combination that could increase prediction accuracy within the corresponding search space. The feature space of GAwAR is denoted by $S$ in Eq.7 where $i_1, i_2, \dots, i_q$ are the indices of the selected features from their corresponding clusters $c_1, c_2, \dots, c_q$ and $({f_1}^s, {f_2}^s, \dots, {f_q}^s)$ is exactly the set of $q$ number of features that GA will use for exploitation and exploration recalling that we stabilized q=100 in this research. Selecting the top 5% through DMC and setting q=100 proves effective for high-dimensional datasets comprising more than 2000 features. However, for datasets with fewer features, customization of these settings becomes essential to ensure the method's applicability in scenarios with lower dimensionality.

$$S = {c_1}_{i_1}, {c_2}_{i_2}, \dots, {c_q}_{i_q} = \left( {f_1}^s, {f_2}^s, \dots, {f_q}^s \right) \tag{7}$$

4.3 GA with adaptive rates

Now that the feature space of the genetic algorithm is specified, the proposed GA with Adaptive Rates (GAwAR) will be applied to find the best combination of 10 features in set $S$, aiming to maximize prediction accuracy. In other words, GA formulates the feature selection problem as a maximization optimization task, with the fitness function being the prediction accuracy. Consequently, the length of the final selected set of features in our proposal in this paper is always 10. Furthermore, the primary contribution of the proposed GAwAR lies in its specific adaptive crossover and mutation rates. GAwAR initializes the probability rates of crossover and mutation, denoted as $P_c$ and $P_m$ respectively. In the event of accuracy improvement plateauing for a consecutive number of iterations, GAwAR responds by considering an increase in the mutation rate to encourage exploration and escape local optima. Simultaneously, it contemplates a decrease in the crossover rate with the same rate. At any given iteration, if the accuracy surpasses that of the previous iteration, the rates revert to their initial values. This dual adjustment is designed to address stagnation and enhance the algorithm's ability to explore diverse solutions by balancing between exploitation (crossover) and exploration (mutation). It's worth noting that the number of individuals in each population iteration remains fixed. Figure 3 illustrates the pipeline for GAwAR, highlighting that the GA operates within a problem-specific feature space (designated as S in Eq.7), known for its diversity and richness conducive to both exploitation and exploration. Additionally, GAwAR employs a developed adaptive crossover and mutation rate (highlighted in red in Figure 3) to effectively balance between exploitation and exploration.

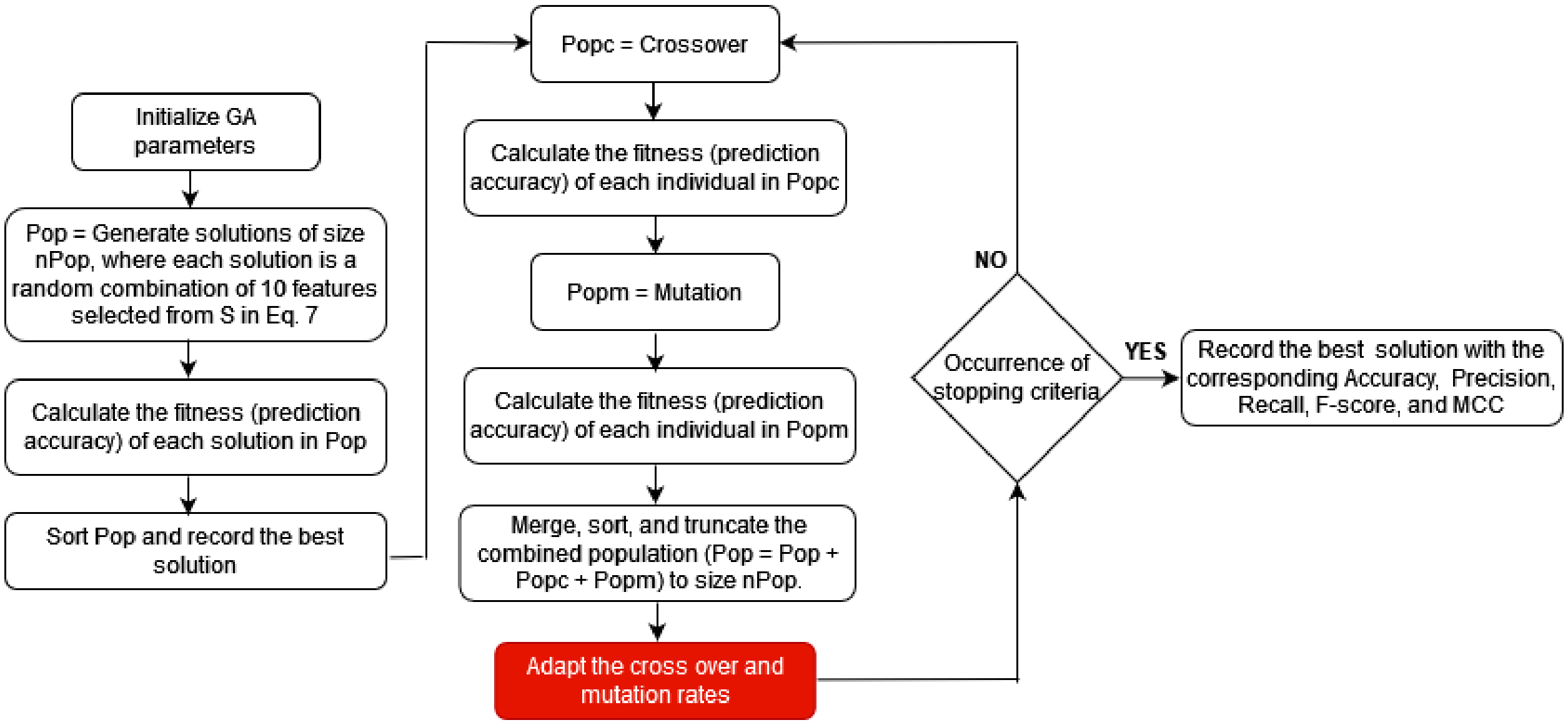

Figure 3. Pipeline for GAwAR

GAwAR utilizes single-point crossover. Assuming two parent individuals of $P_1$ and $P_2$, and a crossover point $C$ randomly selected between 1 and the length of the individuals which we denote by $nVar$. Single-point crossover generates the two offspring individuals $O_1$ and $O_2$ by combining genes of the parents. In Eq. 8 the main body of GAwAR is tasked with generating $n_c = 2 \times \left[ P_c \times {}^{nPop}/_{2} \right]$ number of offsprings ($Pop_c$) in where $P_c$ is adaptively determined in this research.

$$\begin{cases} O_1 = P_1[:C] + P_2[C:] \\ O_2 = P_2[:C] + P_1[C:] \end{cases} \tag{8}$$

The selection of individuals for the crossover operator is performed using Roulette Wheel Selection (RWS) as described in Eq. 11, where the selection probability of each individual is calculated using fitness ratio outlined in Eq. 9. In RWS, a random number $U$ is generated within the range [0, 1], and the selected individual index $k$ is determined using the cumulative probability $W_k$ (Eq.10) at index $k$ according to Eq. 11.

$$P_i = \frac{F_i}{\sum_{j=1}^{nPop} F_j} \tag{9}$$

$$W_i = \sum_{j=1}^{i} P_j \tag{10}$$

$$k = \min\{i: U \ \leq W_i\} \tag{11}$$

Like crossover, the main body of the GA with adaptive rate is tasked with generating $n_m = \lceil P_m \ \times nPop \rceil$ mutants using Algorithm 1 so that $P_m$ is adaptively determined. The input arguments are the parent individual ($x$) and the list of available features for mutation ($seq$). The list of available features for mutation (seq) is calculated by eliminating the feature of $x$ from the feature space ($S$). The output argument of Algorithm 1 is the generated mutant ($y$).

| Algorithm 1. Mutation |
|---|
| **Input**: $x$, $seq$ |
| **Output**: $y$ |
| $R$ = Randomly select a feature from the list $seq$ |
| nVar=length($x$) |
| $J$= randint(1,nVar) |
| $y[J]$=$R$ |
| Return $y$ |

As mentioned previously, the probabilities of the crossover operator ($P_c$) and the mutation operator ($P_m$) are adaptive. According to Algorithm 2, $P_c$ and $P_m$ are initially set to 0.9 and 0.4, respectively. At the end of each iteration, GAwAR investigates the possibility of adapting these probabilities. If the fitness does not improve within 5 consecutive iterations, GAwAR decreases $P_c$ and increases $P_m$ with the same rate that $P_c$ increased. Thus, the number of individuals generated from crossover and mutation remains fixed. This dual adjustment aims to encourage more exploration to prevent the algorithm from getting stuck in local optima and to avoid premature convergence. Additionally, the adaptation constraints force $P_c$ and $P_m$ to stay within the lower and upper bounds of 0 and 1. This strategy allows GAwAR to aggressively explore the search space to potentially discover better solutions than any previously found solution. If, at any iteration, an improvement in the fitness function occurs, $P_c$ and $P_m$ are reverted to their initial values. The total number of iterations is set to 150; nevertheless, if there is no improvement, even with adaptive values for $P_c$ and $P_m$, after reaching 20% of the maximum number of iterations (30 iterations), GAwAR terminates the optimization process. In the last 30 iterations, during which GAwAR evaluates the conditions for early termination of optimization, the values of the ordered pairs of ($P_c$, $P_m$) = (0.9,0.4), (0.7,0.6), (0.5, 0.8), (0.3,1), (0,1),(0,1) as shown in

Figure 4. This indicates that in the last 10 iterations, the individuals generated in each iteration are exclusively mutants, aiming to explore the search space extensively to discover the potential for fitness improvement. The variables Tag and ATag in Algorithm 2 govern the termination conditions for GAwAR: Tag controls the total number of iterations (set to 30) before termination in the absence of improvement, while ATag determines the number of iterations (set to 5) required for adapting crossover and mutation rates. Algorithm 2 represents a section within the main body of the program executed at the conclusion of each iteration, just before incrementing the iteration number.

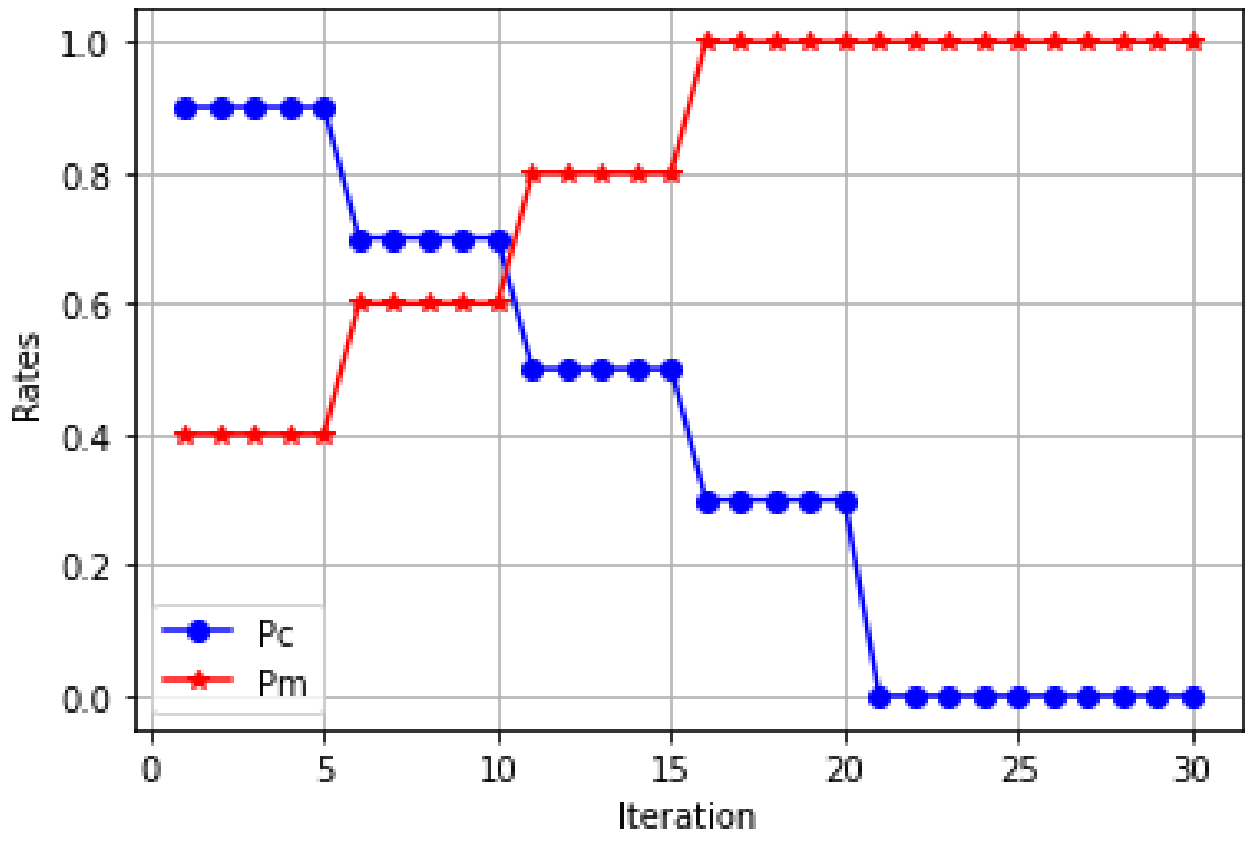

Figure 4. Crossover and mutation rates in the last 30 iterations. Crossover is deactivated by GAwAR starting from iteration 20.

**Algorithm 2. Adaptive rates control**

if (BestFits[it] == BestFits[it - 1]):<br>
    Tag = Tag + 1;   ATag = ATag + 1<br>
else:<br>
    Tag = 1;   ATag = 1<br>
    Revert $P_c$ and $P_m$ to their initial values<br>
if (ATag >5 ):<br>
    ATag = 1;   $P_c = P_c - 0.2$<br>
    if ($P_c < 0.3$):<br>
        $P_c = 0.3$<br>
    $P_m = P_m + 0.2$<br>
    if ($P_m <= 1$):<br>
        $n_m = \lceil P_m \times nPop \rceil$ ; $n_c = 2 \times \left\lceil P_c \times {nPop}/{2} \right\rceil$<br>
    else:<br>
        $P_c = 0$; $n_c = 0$; $n_m = nPop$

## 5. Results

This section begins by introducing the 5 benchmark binary datasets in Section 5.1. Following that, the evaluation criteria and the corresponding hardware/software setup are introduced in Sections 5.2 and 5.3, respectively. Finally, Section 5.4 presents the experimental results and their respective discussions.

### 5.1 Datasets

The binary datasets used in this paper are shown in Table 3. These datasets are extensively used in the literature [11]as benchmark datasets to evaluate the outcome of the proposed feature selection methods. Colon incorporates observations with positive or negative colon cancer results. Central Nervous System (CNS) is a dataset of two classes of survivor (alive after treatment) and failure (dead after treatment) samples. GLI contains gliomas of two categories. The gene expression data SMK are obtained from smokers with lung cancer and smokers without lung cancer. Leukemia contains bone marrow samples of 2 specific categories. The sample and feature sizes vary from 60 to 187 and 2000 to 22283 which clearly show the high dimensionality. All datasets listed in Table 3 exhibit evident imbalances, with the exception of SMK.

Table 3. Datasets under evaluation

| Binary dataset | Sample size | Feature size | Sample distribution |
|---|---|---|---|
| Colon | 62 | 2000 | 22-40 |
| CNS | 60 | 7129 | 21-39 |
| GLI | 85 | 22283 | 26-59 |
| SMK | 187 | 19993 | 90-97 |
| Leukemia | 72 | 7129 | 47-25 |

### 5.2 Evaluation criteria

The evaluation criteria in this paper include overall accuracy, balanced accuracy, precision, recall, F-score, and Matthews Correlation Coefficient (MCC). It is important to note that the subset length of the selected set of features is consistently set to 10, as indicated in Figure 1. The imbalanced distribution of samples in datasets leads to a bias towards the majority class. In such cases, balanced accuracy which provides equal weight to all the classes in a dataset might better reflect the imbalanced. This is why we also investigate the balanced accuracy besides the overall accuracy. The accuracies can be calculated by Eqs.(12-13) using recall and specificity in Eqs(14-15).

$$Overall\ accuracy = \frac{TP+TN}{TP+TN+FP+FN} \qquad (12)$$

$$Balanced\ accuracy = \frac{Recall+Specificity}{2} \qquad (13)$$

Where

$$Recall = \frac{TP}{FP+FN} \qquad (14) \qquad\qquad Specificity = \frac{TN}{TN+FP} \qquad (15)$$

In addition to recall, precision and F-score, are crucial for evaluating prediction accuracy in situations of class imbalance as shown in Eqs. (16-17). Moreover, the Matthews Correlation Coefficient (MCC), denoted in Eq. (18), offers a more comprehensive measure of classification accuracy. The MCC considers both false positives and false negatives, making it particularly useful when dealing with imbalanced datasets or uneven class representation. The MCC is scaled from -1 to 1, where 1 indicates perfect prediction, 0 suggests random prediction, and -1 implies complete disagreement between the prediction and observation. The equations represented by Eqs. (12-18) can be computed using the True Positive (TP), True Negative (TN), False Positive (FP), and False Negative (FN) values obtained from the confusion matrix.

$$Precision = \frac{TP}{TP + FP} \tag{16}$$

$$Fscore = 2 \times \frac{Precision \times Recall}{Precision + Recall} \tag{17}$$

$$MCC = \frac{TP \times TN - FP \times FN}{\sqrt{(TP+FP)(TP+FN)(TN+FP)(TN+FN)}} \tag{18}$$

5.3 Experimental configuration

This paper employs a Decision Tree (DT) as a classifier for evaluating measurement criteria. The default configuration from the Python library is utilized. Consequently, the DT in this study employs the Gini index to compute impurity, and default values are applied to parameters such as random_state, class_weight, and ccp_alpha. The stratified train-test split is used with test size=20% and the accuracy of a selected subset is the average of 10 times stratified train-test split. This research is implemented using Python 3.9.13 platform on a computer with a Core i5 processor (1.60 GHz–2.30 GHz), 12 GB RAM, a 720 GB HDD, and 64-bit Windows 10 Pro operating system.

5.4 Experimental results

Table 4 presents the average overall and balanced accuracy before and after the application of DMC-GAwAR across 5 runs. Table 4 clearly illustrates a significant improvement in dataset accuracy. Notably, in certain datasets like CNS and GLI, balanced accuracy is slightly lower. This discrepancy is expected, as overall accuracy can be biased by the majority class. However, it is noteworthy from Table 4 that the difference in accuracy between overall and balanced accuracy is not statistically significant generally. Therefore, for the subsequent figures and tables in the paper, we have chosen to adhere to overall accuracy.

Figure 5 illustrates that the proposed method, DMC-GAwAR, notably enhances precision, recall, F-score, and MCC across all datasets. This improvement is particularly striking in leukemia, where all measures reach perfection at 1. In medical datasets, precision and recall hold substantial importance. Precision becomes critical when the cost of false positives is high—such as in diagnostic tests, aiming to minimize unnecessary treatments or alarms. Conversely, recall is crucial when the cost of false negatives is significant. Missing a true positive (false negative) in medical diagnoses could delay necessary treatments or interventions. Figure 5 clearly demonstrates that DMC-GAwAR significantly elevates precision and recall, metrics crucial for evaluating the performance of a classification or information retrieval system, in addition to overall accuracy.

### 5.4.1 DMC Vs. Binary frequency-based rankers

Figure 6 illustrates that it's challenging to definitively determine the superior ranker since each demonstrates optimal performance with specific datasets. For instance, in Colon, DMC appears to excel, yet it exhibits the poorest performance in SMK. Conversely, SLI-$\gamma$, which performs poorly in Colon and GLI, showcases near-optimal performance in SMK. In Leukemia, all rankers demonstrate high competitiveness. These findings highlight the nuanced nature of datasets, varying in shapes such as concave, convex, linear, nonlinear, and more.

Table 4. Average accuracy (overall and balanced) before and after the application of DMC-GAwAR with a subset length of 10

| Dataset | Overall accuracy before | Balanced accuracy before | Overall accuracy after | Balanced accuracy after |
|---|---|---|---|---|
| Colon | 0.74 | 0.74 | 0.96 | 0.96 |
| CNS | 0.58 | 0.52 | 0.91 | 0.86 |
| GLI | 0.79 | 0.76 | 0.96 | 0.94 |
| SMK | 0.59 | 0.59 | 0.80 | 0.79 |
| Leukemia | 0.84 | 0.84 | 1 | 1 |

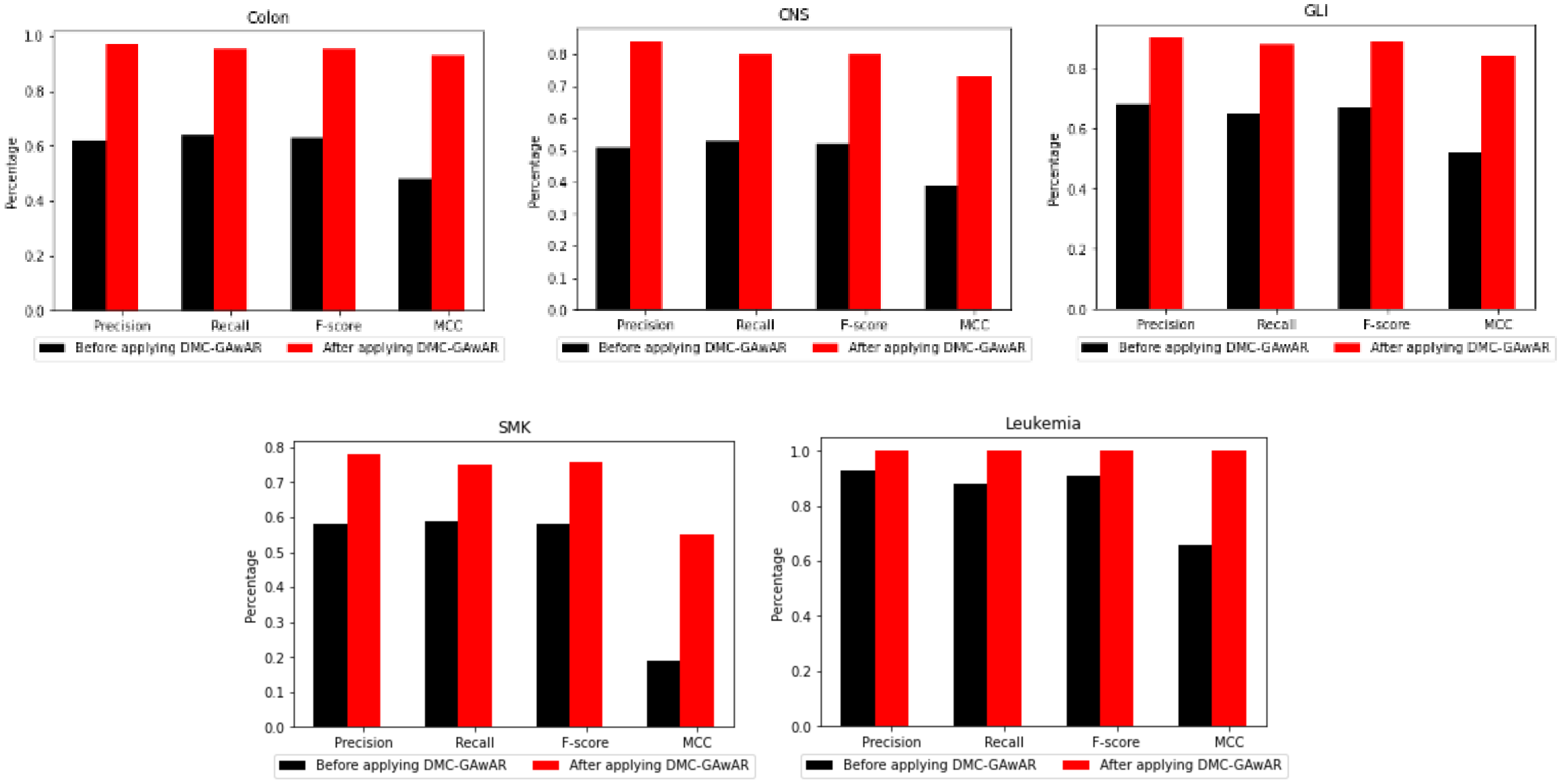

Figure 5. Precision, recall, F-score, and MCC before and after DMC-GAwAR application.

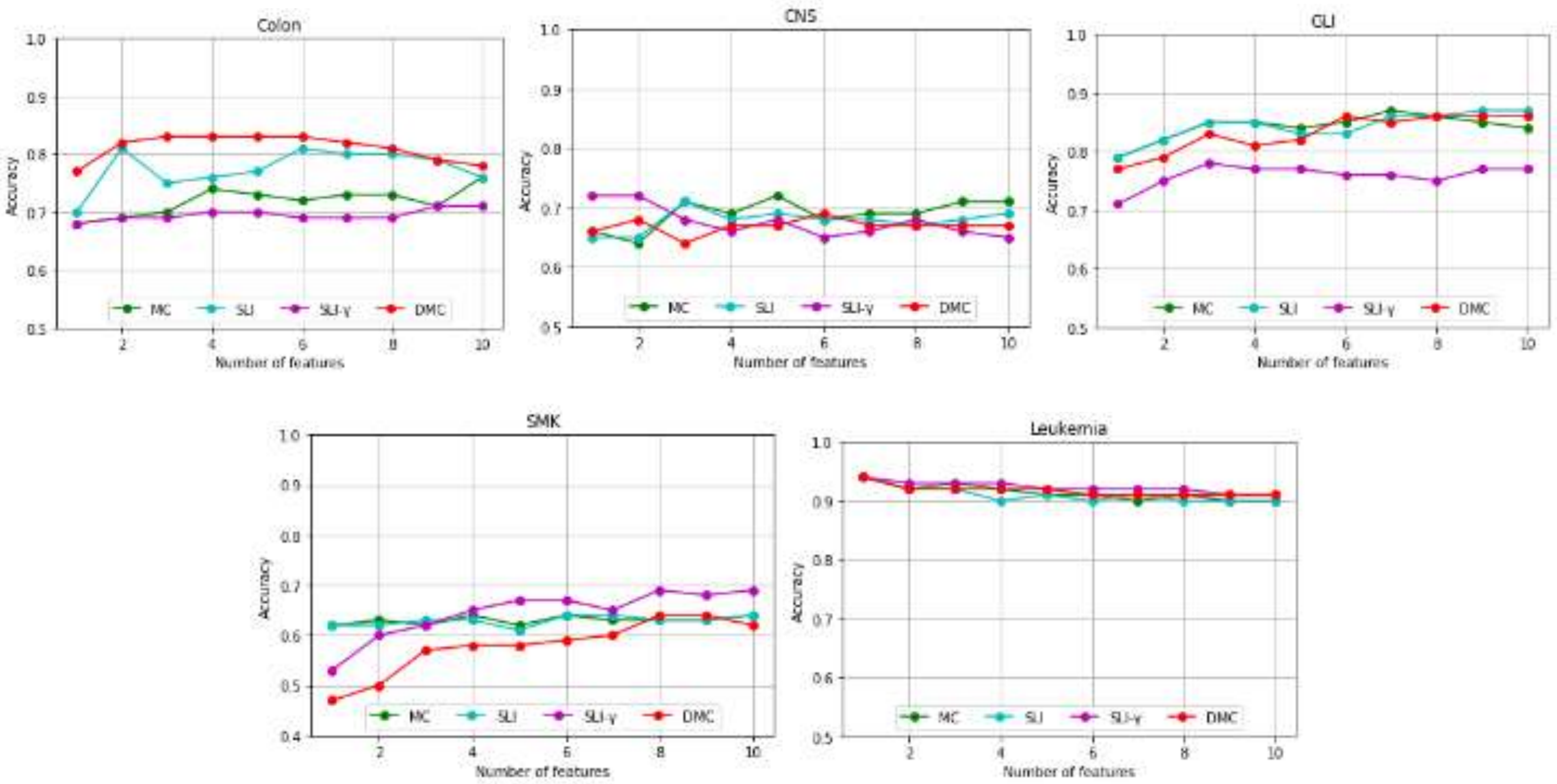

Figure 6. Comparison of DMC with existing binary frequency-based rankers

Moreover, it's crucial to acknowledge that these rankers typically select distinct features due to their intrinsic formulations. For instance, in the colon dataset, only 40% of the top 10 features and 46% of the top 100 features chosen by MC and DMC are similar. This disparity underscores how these two methods consistently identify distinct sets of features. Given that rankers assess features individually without accounting for feature combinations, the outcomes suggest two significant considerations:

1. Conducting experimental investigations on each ranker per dataset is crucial, indicating the absence of a globally superior ranker applicable across all datasets.
2. Exploring how combinations of features impact predictions could involve using wrapper feature selections. However, to mitigate multicollinearity and potential noise among features, an effective strategy might entail employing a wrapper approach on the best features identified by a ranker.

The time complexity of DMC, expressed in big O notation, is $O(mn\,log\,n)$, where $m$ represents the number of features, and $n$ is the number of samples in the dataset. This analysis assumes that sorting is performed using merge sort, with a time complexity of $O(n\,log\,n)$. As indicated in Figure 2, the process of identifying the mutual congestion region (the interference region where labels are not clearly separable) and the corresponding calculations in DMC take $O(n)$ time. Consequently, we can infer that DMC incurs an equivalent time cost as MC, SLI, SLI_γ, and MPR, and is faster than EMC.

### 5.4.2 Convergence graph of GAwAR

Figure 7 displays the convergence graph of the GAwAR, seemingly confirming the smooth maximization of fitness. This pattern indicates a consistent and progressive enhancement in solutions discovered by the algorithm, showcasing its effectiveness in exploiting and exploring the search space. Furthermore, the smooth rise in fitness implies that the maximization process isn't purely random. Moreover, by clustering the top 5% of features identified by DMC into 100 clusters and selecting one feature from each cluster to mitigate potential multicollinearity, the feature space for the proposed GAwAR comprises 100 features. The objective of this research is to identify the optimal combination of 10 features that maximizes prediction accuracy through the proposed GAwAR. Consequently, the combination of $\binom{100}{10} = \frac{100!}{10!(100-10)!}$ equals 17,310,309,456,440, signifying the vast array of possible ways to select 10 features from a pool of 100 where order is insignificant. This expansive search space emphasizes the challenge at hand, affirming the use of GA, which efficiently approximates optimal solutions through limited evaluations. The graphs also incorporate blue circles and red rectangles, indicating locations where the probability of crossover and mutation dynamically adapts in GAwAR. Notably, the frequency of adaptive changes varies across datasets, with SMK exhibiting the highest adaptability over numerous iterations. The red rectangles signify points where the mutation rate reaches 100%, indicating that GAwAR exclusively employs mutation for exploration (all generated solutions originate from the mutation operator). It is evident that in the Colon and SMK datasets, employing maximum exploration, as indicated by these red rectangles, leads to improved solutions, particularly notable after the second red rectangle where optimization occurs. In all cases, GAwAR terminates if, after 5 consecutive adaptations (30 consecutive iterations), no optimization is observed.

While Figure 7 illustrates the number of adaptations in crossover and mutation rates for each dataset, Figure 8 provides a detailed account of these adaptations by plotting the mutation rate changes (the crossover changes will have an opposite trend). For instance, in the 14th iteration of the Colon dataset, the first adaptation occurs with the mutation rate increasing from 0.4 to 0.6. However, by the 17th iteration, the mutation rate reverts to 0.4, a phenomenon triggered by an optimization in the fitness function, which prompts GAwAR to reset all rates to their respective initial values. Subsequently, at the 22nd iteration, the mutation rate rises again to 0.6. However, this increase fails to contribute to fitness optimization. Consequently, by the 27th iteration, the mutation rate escalates to 0.8, reaching 1 at the 32nd iteration. GAwAR maintains a mutation rate of 1 and a crossover rate of 0.3 for the next 5 consecutive iterations (iterations 32-36). Starting from the 37th iteration, GAwAR exclusively generates solutions through mutation, deactivating crossover until the 45th iteration, wherein the algorithm achieves an increase in the fitness function, attributed to perfect exploration. As a result, the mutation rate reverts to its initial value of 0.4, and correspondingly, the crossover rate returns to 0.9. This cycle continues until the GAwAR concludes at the 76th iteration. The red thick lines on the graphs in Figure 8 indicate iterations characterized by perfect exploration, where GAwAR solely utilizes its mutation operator for optimization and completely deactivates its crossover operator.

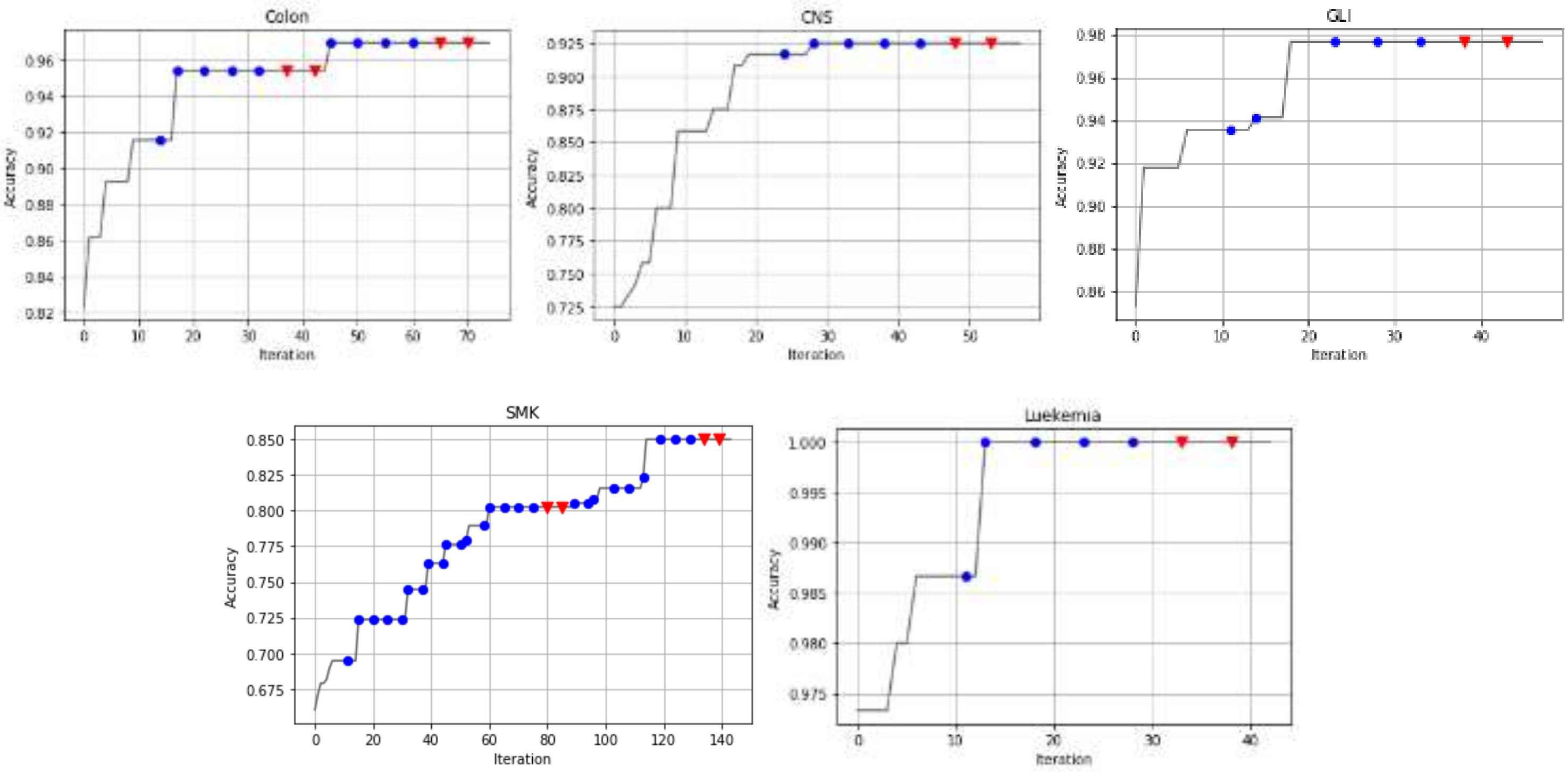

Figure 7. Convergence graph of GAwAR incorporating adaptation rates' locations.

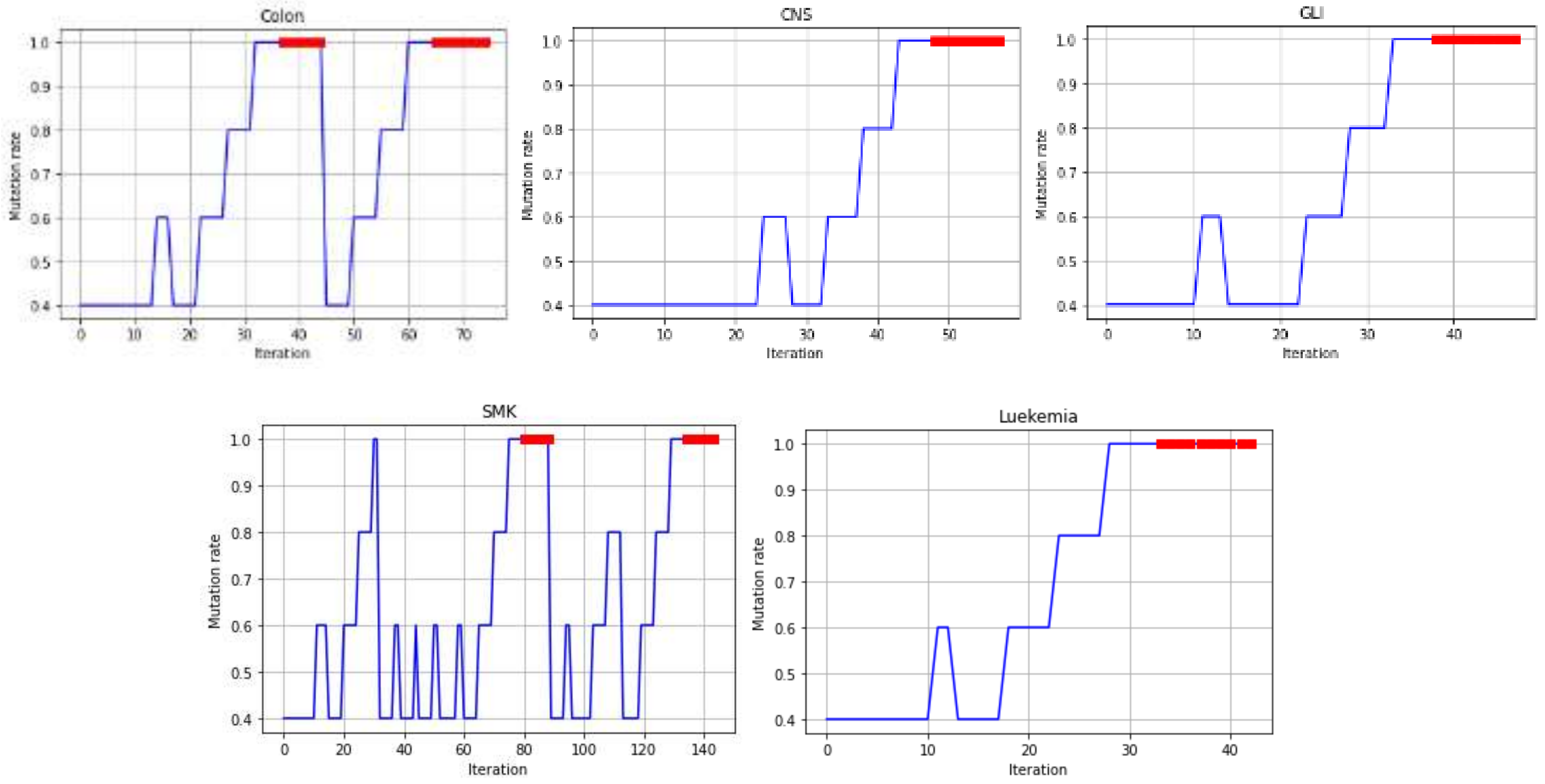

Figure 8. Plots illustrating the change in mutation rate across iteration.

Additionally, Figure 9 (right side) compares the accuracy between DMC-Random Selection and DMC-GAwAR. This comparison distinctly demonstrates that GAwAR, functioning as a guided random search, outperforms a purely random selection. The accuracy of DMC-Random Selection involves the random selection of 10 features from a pool of 100. The accuracy values shown in Figure 9 represent the average across 3 runs. Evidently, the average accuracy of DMC-Random Selection falls below the initial accuracy (depicted by horizontal blue lines, indicating accuracy without feature selection) in the Colon, CNS, and GLI datasets. Furthermore, there is a considerable accuracy difference between DMC-Random Selection and DMC-GAwAR across all datasets. This discrepancy provides further support for the efficacy of GA within the wrapper feature selection process. The average Number of Function Evaluations (NFE) in 5 runs of GA are depicted in Figure 9 (left side). This observation confirms that GAwAR effectively approximates the global maximum while utilizing considerably fewer solutions (2173, 1926, 1221, 1788, 1199 for Colon, CNS, GLI, SMK, and Leukemia respectively) compared to the search space of $\binom{100}{10}$.

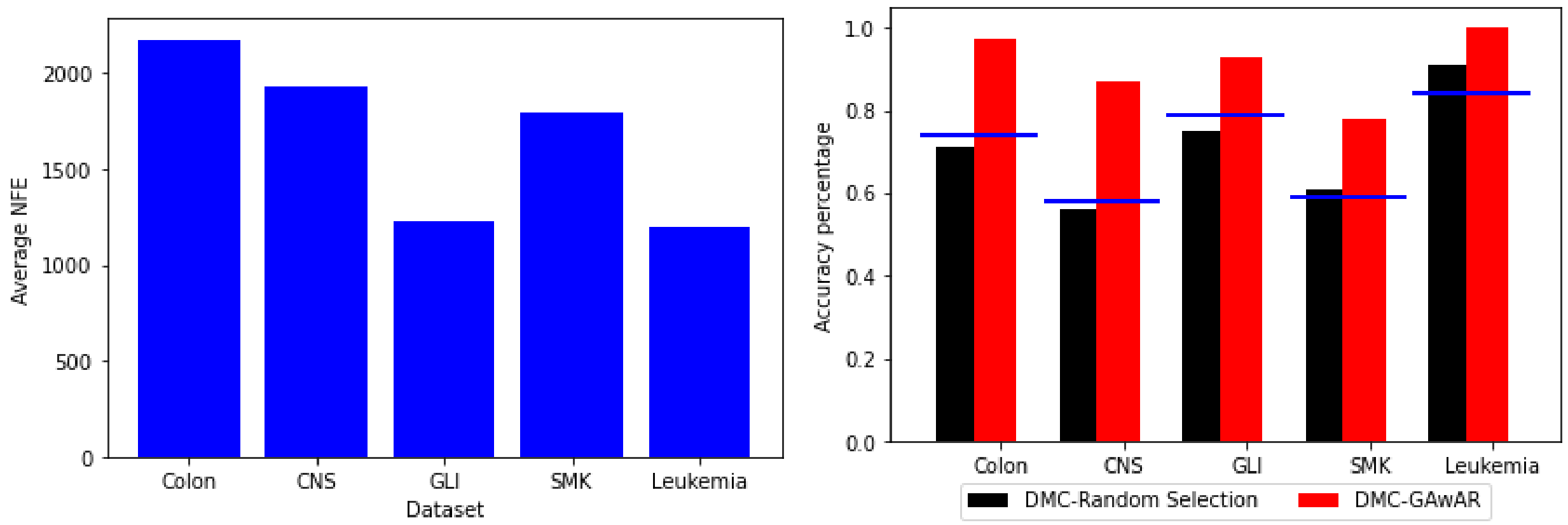

Figure 9. The average NFE of GAwAR for datasets (left side) and the comparison of the average overall accuracy between DMC-Random Selection against DMC-GAwAR (right side)

### 5.4.3 Comparative analysis

Table 5 compares the average Subset Length (SL) and Overall Accuracy (Acc) across 5 runs for each method. It is evident that, on average, DMC-GAwAR attains higher accuracy, except in the case of EIT-bBOA, where EIT-bBOA marginally outperforms it. WOA-MC and PEFS are hybrids of 2 filters and an ensemble of 4 filters, respectively. Notably, EIT-bBOA, EMC-DWES, MPR-MDES, and the proposed DMC-GAwAR incorporate wrapper feature selection, which may be a key factor contributing to the superior accuracies observed in comparison to WOA-MC and PEFS. Among the methods listed in Table 5, only WOA-MC and DMC-GAwAR are exclusively designed for binary classification datasets, while the remaining are applicable to both binary and multiclass datasets. Therefore, Table 5 substantiates that when working with a binary dataset, DMC-GAwAR excels in capturing high dimensionality by identifying informative features.

Table 5. Comparison of DMC-GAwAR (last column) with state-of-the-art methods. SL denotes average subset length, and Acc represents mean accuracy calculated by decision tree.

| Dataset | WOA-MC [8] | | PEFS [15] | | EIT-bBOA [16] | | EMC-DWES [11] | | MPR-MDES [12] | | DMC-GAwAR | |
|---|---|---|---|---|---|---|---|---|---|---|---|---|
| | SL | Acc | SL | Acc | SL | Acc | SL | Acc | SL | Acc | SL | Acc |
| Colon | 10 | 0.74 | 40 | 0.76 | 30 | 0.86 | 6 | 0.91 | 6 | 0.94 | 10 | **0.96** |
| CNS | 10 | 0.72 | 20 | 0.66 | 30 | 0.84 | 26 | 0.82 | 6 | 0.88 | 10 | **0.91** |
| GLI | 10 | 0.86 | 100 | 0.85 | 30 | 0.84 | 29 | 0.91 | 5 | 0.93 | 10 | **0.96** |
| SMK | 10 | 0.66 | 80 | 0.71 | 30 | **0.82** | 33 | 0.70 | 6 | 0.75 | 10 | 0.80 |
| Leukemia | 10 | 0.92 | 10 | 0.94 | 30 | 0.89 | 17 | 0.97 | 3 | 0.99 | 10 | **1** |

## 6. Conclusion

This paper presents a new frequency-based ranker applicable to binary classification dataset namely Distance-based Mutual Congestion (DMC) and combines it with a wrapper feature selection Genetic Algorithm with Adaptive Rates (GAwAR). The top 5% features are selected by DMC and clustered into 100 clusters to generate the feature space with the size of 100. Then, GAwAR approximates the optimal solution that contains 10 features from a search space that contains $\binom{100}{10}$ solutions. Additionally, GAwAR exploits an intelligent algorithm for adaptation of crossover and mutation rates which assists it to escape from premature convergence. The proposed hybrid DMC-GAwAR approximates the combination of best 10 features that could increase the classifier accuracy (decision tree in our paper). The paper substantives how DMC-GAwAR could effectively increase the accuracy in binary datasets in comparison with some state-of-the-art works.

DMC assumes a linear distribution of data, similar to its counterparts, and consequently, it may struggle to identify informative features with non-linear distribution values. Therefore, a key future direction for this research involves developing a frequency-based ranker capable of recognizing informative features characterized by non-linear distributions with applicability on multiclass data.